\documentclass[runningheads]{llncs}

\usepackage[T1]{fontenc}
\usepackage{graphicx,longtable,multicol}

\begin{document}

\title{Thematic and Task-Based Categorization of K-12 GenAI Usages with Hierarchical Topic Modeling\thanks{Accepted at the International Conference on Computer-Human Interaction Research and Applications (CHIRA), 2025;  Cite as: Schneider, J., Hasler, B. S., Varrone, M., Hoya, F., Schroffenegger, T., Mah, D.-K., and Peböck, K. (2025). Thematic and task-based categorization of K-12 GenAI usages with hierarchical topic modeling. International Conference on Computer-Human Interaction Research and Applications (CHIRA).}}

\author{
  Johannes Schneider\inst{1} \and
  Béatrice S. Hasler\inst{1} \and
  Michaela Varrone\inst{3} \and
  Fabian Hoya\inst{2} \and
  Thomas Schroffenegger\inst{2} \and
  Dana-Kristin Mah\inst{4} \and
  Karl Peböck\inst{2}
}

\institute{
  University of Liechtenstein, Vaduz, Liechtenstein\\
  \email{\{johannes.schneider, beatrice.hasler\}@uni.li}\\[0.5ex]
  \and
  University College of Teacher Education Vorarlberg, Feldkirch, Austria\\
  \email{\{fabian.hoya, thomas.schroffenegger, karl.peboeck\}@ph-vorarlberg.ac.at}\\[0.5ex]
  \and
  School Authority Liechtenstein, Vaduz, Liechtenstein\\
  \email{michaela.varrone@llv.li}\\[0.5ex]
  \and
  Leuphana University of Lüneburg, Lüneburg, Germany\\
  \email{dana-kristin.mah@leuphana.de}
}

\authorrunning{J. Schneider et al.}


\maketitle

\begin{abstract}
We analyze anonymous interaction data of minors in class-rooms spanning several months, schools, and subjects employing a novel, simple topic modeling approach. Specifically, we categorize more than 17,000 messages generated by students, teachers, and ChatGPT in two dimensions: content (such as nature and people) and tasks (such as writing and explaining). Our hierarchical categorization done separately for each dimension includes exemplary prompts, and provides both a high-level overview as well as tangible insights. Prior works mostly lack a content or thematic categorization. While task categorizations are more prevalent in education, most have not been supported by real-world data for K-12. In turn, it is not surprising that our analysis yielded a number of novel applications.
In deriving these insights, we found that many of the well-established classical and emerging computational methods, i.e., topic modeling, for analysis of large amounts of texts underperform, leading us to directly apply state-of-the-art LLMs with adequate pre-processing to achieve hierarchical topic structures with better human alignment through explicit instructions than prior approaches.
Our findings support fellow researchers, teachers and students in enriching the usage of GenAI, while our discussion also highlights a number of concerns and open questions for future research.
\end{abstract}

\section{Introduction}
Generative artificial intelligence (GenAI) has rapidly emerged as a transformative tool across diverse domains, and education is no exception. From adaptive tutoring systems to automated essay scoring\cite{min23,sch24grad}, proposals for GenAI–driven interventions promise to personalize learning, reduce teacher workload, and foster student engagement. Yet despite this growing enthusiasm, empirical evidence on how learners—particularly minors—actually interact with GenAI in real-world educational settings remains scarce. Most prior work relies on simulations in lab-settings (e.g., \cite{sch24val}), self-reported surveys (e.g., \cite{lang24}), or small-scale pilots (e.g., \cite{Ng2024,Fokides2025}), leaving a critical gap between theoretical potential and classroom practice.

In this study, we bridge that gap by conducting an in-depth analysis of over 17,000 anonymized messages exchanged between students and ChatGPT across 11 months of classroom activities spanning multiple subjects, including mathematics, literature, and science. By examining authentic interaction logs from primary and secondary schools, we uncover the ways in which minors and teachers leverage GenAI in an education setting.

Our contributions are threefold. The first two are related to exploring and structuring GenAI usages in K-12 education, grounded in empirical data rather than hypothetical scenarios. We present a systematic taxonomy of GenAI (i) prompt content and (ii) prompt tasks, i.e., tasks to be addressed by GenAI. Previous studies have categorized generative AI usage by user motivations or task types rather than by content themes\cite{skj24,hug24,ama25}. Our content-based categorization enables educators to identify which topics students interact with most frequently, helping them align instructional resources accordingly. A content-oriented taxonomy also  provides deeper insights into students' genuine interests, thereby informing curriculum designers about which areas require greater support or enrichment. 
Our task categorization serves a dual purpose: On the one hand, it confirms existing usages, while on the other hand, it also points to novel, emerging applications not yet documented in the academic literature. While GenAI has many benefits, we also provide a critical reflection on its use in K-12 education that also points to future research directions.

The third contribution relates to methodology: our work provides interesting insights into the integration of GenAI as a computational method for topic modeling. Topic modeling allows to identify key themes in diverse sets of large amounts of documents, ranging from long books to short text messages. It is the only practically feasible approach for large amounts of texts and has been used extensively across various areas of research. For instance, one of the leading methods - Latent Dirichlet Allocation (LDA) \cite{blei2003latent} - comes with more than 50,000 citations. However, these classical topic models, though still being in use, have a number of shortcomings \cite{schn18}. Newer approaches rely on complex algorithms leveraging LLMs. But they maintain fundamental issues such as poor customizability and lack of hierarchical structuring\cite{pham24,sch24top,wang23}. These shortcomings lead to unstructured topics that are often not aligned with human intentions and hard to grasp. In our work, we perform an assessment and describe an approach to leveraging state-of-the-art LLMs for topic modeling to overcome shortcomings in prior topic models, providing valuable insights on how to structure large amounts of data, while acknowledging also challenges in doing so.

\section{Methodology}
Our findings rely on real-world interaction data with ChatGPT, which was processed and analyzed by researchers with the help of computational methods.

\subsection{Data Collection and Pre-processing}
Data stemmed from a small German-speaking European country that launched a pilot study using generative AI in K-12 across the entire country. They relied on ChatGPT-4o as a model for students and teachers. However, students and teachers were not accessing ChatGPT through the standard web-interface but rather through special software leveraging OpenAI's API and providing a number of helpful management functionality such as creating virtual classrooms. Students were either explicitly instructed to address a task using GenAI or could use it on their own account. We did not have access to any private students' or teachers' accounts that they might use outside of the class-room setting. We also verified that student usage occurred almost exclusively during school hours, e.g., not on weekends or at night. That is, student usage was confined to classroom interactions. Teachers also used ChatGPT for preparation and planning. All students were instructed on the usage of ChatGPT (prior to our study). Data were only collected and stored anonymized to protect privacy, adhering to standard ethical guidelines for research involving minors \cite{Cohen2018,BERA2018}. That is, each row in our data file consisted of a conversation ID, role (user / AI), message and timestamp, but no information allowing to identify individuals. 

\begin{figure}
    \centering
    \includegraphics[width=0.75\linewidth]{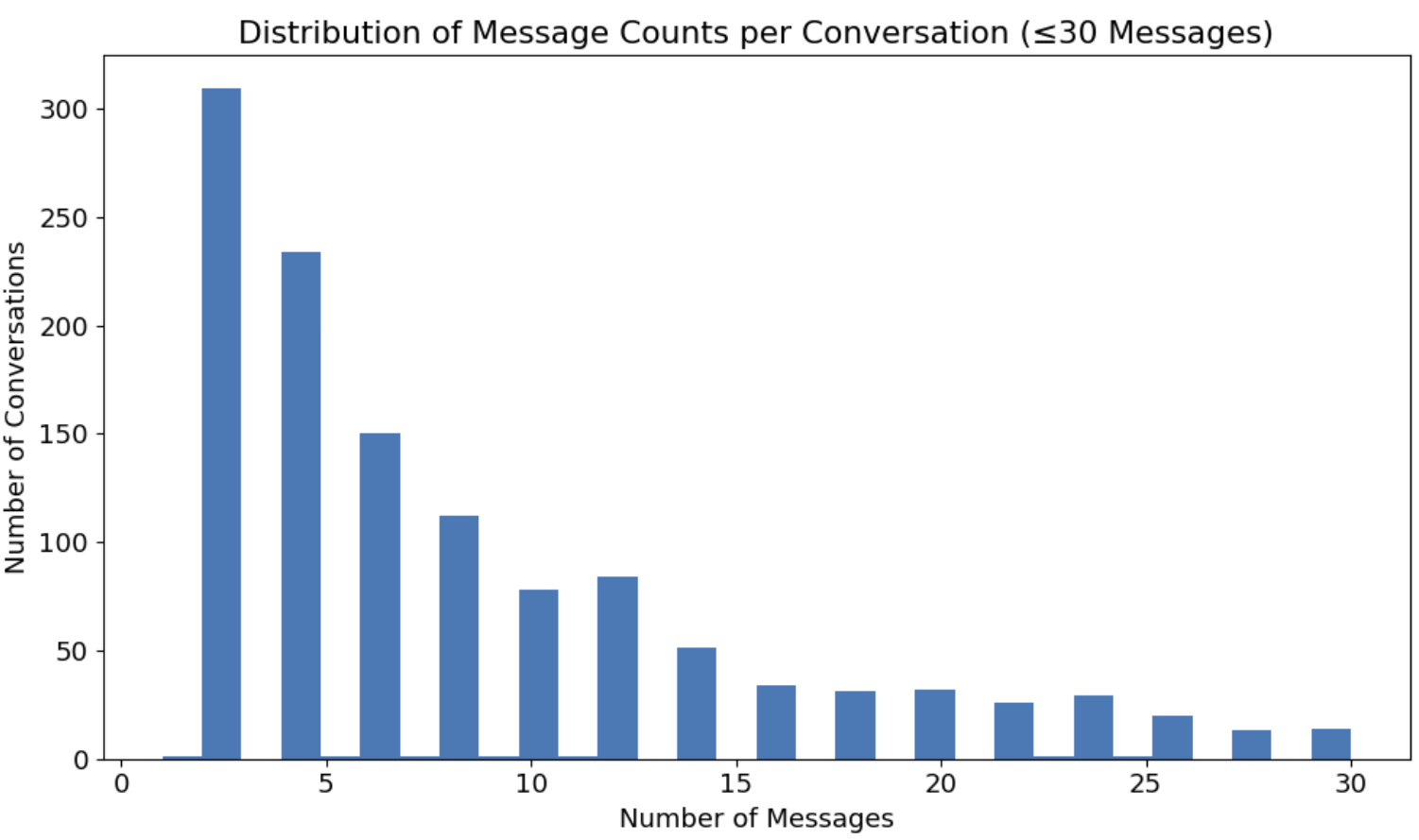}
    \caption{Distribution of conversation frequency by number of messages: Most conversations contain few messages}
    \label{fig:dist}
\end{figure}

Thus, our interaction data is based on anonymized logs of ChatGPT conversations generated by minors over an extended period, spanning from August 2024 to June 2025. The data stems from various schools and different classes with each school covering various age groups within the K-12 spectrum. 
The dataset comprises over 17,294 individual messages spanning multiple academic subjects, including mathematics, literature, and science. These messages can be partitioned into 1,339 conversations, each consisting of at least one user prompt followed by a GenAI response. As shown in Figure \ref{fig:dist} , more than 91\% of all conversations contain at most 30 messages. Notably, we identified five conversations with more than 120 messages (but none surpassing 250), which are likely due to users continuing different tasks within the same chat sessions rather than initiating new ones. 

For presentation in the manuscript, we translated all interaction data to English using ChatGPT 4.5. We asked for concise translations to minimize modifications. Our manual validation using 30 samples, yielded that translations were accurate, but spelling and grammar mistakes, which were common in the original German prompts, were mostly eliminated.

\subsection{Data selection}
We aimed to categorize interactions along two dimensions using natural language processing (NLP): content and subjects being discussed (e.g., animals, places) and tasks being performed (e.g., generating ideas, providing feedback).

To identify the prevalent content and subjects, we focused on nouns as they frequently capture core discussion topics \cite{Jurafsky2019Speech}. The text data was tokenized, and part-of-speech tagging was applied using  the spaCy library \cite{Honnibal2020SpaCy}, enabling efficient and accurate extraction of nouns. As a second approach, we employed a simple heuristic focusing on capitalized words (Nouns are capitalized in German). We found that the second approach yielded better outcomes meaning fewer words misclassified as nouns. Therefore, we relied on it. Subsequently, we identified the 400 most common nouns using frequency analysis. The frequency of the 400 chosen words varied between 3200 to 140. We deemed occurrences of 140 sufficient to cover key themes without getting lost in spurious occurrences of nouns.    

To identify tasks, we focused on the initial prompt of a conversation. That is, the initial prompt typically captures the task the AI should perform. We first removed duplicates leaving 1014 conversations out of 1339. Duplicates commonly emerge from copy\&paste. As GenAI's responses are often not identical for the same prompt, copy\&paste can be viable strategy to obtain different responses if the initial one is not satisfactory. Furthermore, students are also inclined to copy\&paste task descriptions directly into ChatGPT. As some prompts were extremely long - for example, containing entire essays for which students sought feedback - we kept only the first 100 characters of the first prompt in each conversation for analysis. 

\begin{figure}
    \centering
    \includegraphics[width=\linewidth]{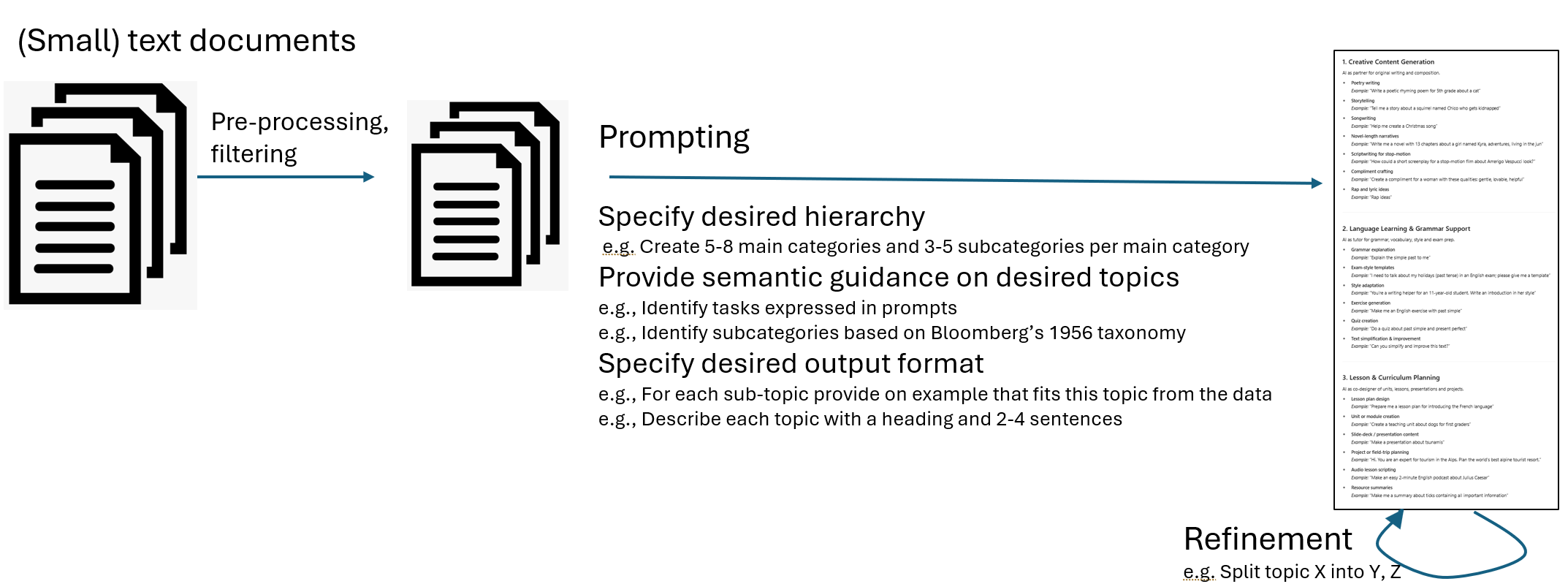}
    \caption{``Simple'' Topic modeling using LLMs ``out-of-the-box'' enables hierarchical, customizable topic generation, which is not possible with prior methods.}
    \label{fig:top}
\end{figure}

As a next step, we identified topics and themes within these data. Scientific approaches can roughly be split into computational and manual approaches. More modern computational approaches for topic modeling increasingly rely on LLMs \cite{sch24top,wang23,pham24}, while older, less powerful models often rely on pure statistical models \cite{blei2003latent}. We found that the native use of state-of-the-art LLMs (we used ChatGPT 4.5 and the state-of-the-art reasoning model o3-pro) gave better outcomes than classical approaches, we have employed. In particular, most specialized models like \cite{sch24top,blei2003latent} struggle to produce hierarchical categorizations. Instead, they tend to produce a large number of topics, some of which lack meaningful coherence and require manual processing, as acknowledged in peer-reviewed methodology papers on topic modeling \cite{debo16} and publications in leading academic journals \cite{hack20}.
The overarching approach is illustrated in Figure \ref{fig:top}. It relies on data pre-processing and prompting often comprising of elements such as desired hierarchy of topics, semantic guidance on desired topics, and output format. Manual refinement might be needed, but we found topics to be of much higher quality and much easier to interpret than for classical approaches.  We elaborate more in the discussion section.

\section{Findings}
We present two complementary classification dimensions: content categories and task categories. Content categories are often associated with educational subjects (e.g., biology, geography), while task categories capture more general forms of learning support that span across subjects (e.g., explanation, feedback, creative content generation). However, we acknowledge that these dimensions are not entirely independent, as specific content areas often shape the nature and framing of tasks.

\subsection{Content Dimension}
\begin{table}[]
    \centering
\caption{Categorization of Interaction Data: Major Themes and Associated Keywords} \label{tab:subj} 
\begin{tabular}[!t]{p{3cm}|p{9cm}}

\textbf{Topic/Subtopic} & \textbf{Words} \\
\hline
\multicolumn{2}{l}{\textbf{1. Education \& Learning}}  \\ \hline
School \& Classes & Student, Teacher, Class, High School, Apprenticeship \\ \hline
Writing \& Language & Text, Words, Sentences, Grammar, Spelling, Language, Sentence, Introduction, Conclusion, Main part \\ \hline
Knowledge \& Information & Information, Questions, Answers, Explanation, Understanding, Details, Topic, Title \\ \hline
Presentation \& Communication & Slide, Presentation, Prompt, Report, Description, Content \\ \hline

\multicolumn{2}{l}{\textbf{2. Nature \& Environment} } \\ \hline
Flora \& Fauna & Cat, Dog, Animals, Plants, Birds, Eagle, Owl, Eagle owl, Squirrel, Wildlife park, Flowers, Trees \\ \hline
Landscape \& Locations & Forest, Village, City, Park, Glacier, Sea, Earth, Sky, Mars, Moon, Stars, Region, Atmosphere, Environment \\ \hline
Weather \& Natural Phenomena & Snow, Water, Rain, Ice, Tsunamis, Weather, Snowflakes, Climate \\ \hline
Ecology \& Sustainability & Nutrition, Fertilizer, Protection, Environment, Ground, Disposal, Bacteria, Health, Skin cancer, Immune system \\ \hline

\multicolumn{2}{l}{\textbf{3. Family \& Relationships} } \\ \hline
Family \& People & Family, Father, Children, Girl, Boy, Friends, Friendship, Relationships, Person, Man, Woman \\ \hline
Celebrations \& Holidays & Christmas, Christmas Story, Santa Claus, Christmas tree, Christmas time, Gifts, Reindeer, Merry \\ \hline
Community \& Social Interaction & Community, Team, Teamwork, Collaboration, Engagement, Appreciation, Respect, Compliment \\ \hline

\multicolumn{2}{l}{\textbf{4. Arts \& Entertainment} } \\ \hline
Literature \& Storytelling & Story, Chapter, Adventure, Narrator, Characters, Plot, Book, Stories, Poem \\ \hline
Film \& Performance & Scene, Film, Screenplay, Actors, Role, Expressions, Mood, Impressions \\ \hline
Games \& Activities & Gaming, Minigolf, Quiz, Riddle, Fun, Activities, Game, Exercises \\ \hline

\multicolumn{2}{l}{\textbf{5. Career \& Professional Development} } \\ \hline
Careers \& Jobs & Career choice, Professions, Profession, Work, Tasks, Responsibility, Company, Apprenticeship, Skills \\ \hline
Personal Growth & Goals, Development, Strengths, Abilities, Interests, Experiences, Career Exhibition, Recognition, Feedback \\ \hline
Education \& Guidance & Tips, Opportunities, Decision, Steps, Target group, Presentation, Information, Future, Questions \\ \hline

\multicolumn{2}{l}{\textbf{6. Health \& Wellbeing} } \\ \hline
Body \& Health & Skin, Body, Acne, Symptoms, Treatment, Health, Immune system, Protection, Causes \\ \hline
Feelings \& Emotions & Feelings, Joy, Love, Fun, Smile, Mood, Courage, Patience, Attitude \\ \hline
Lifestyle \& Activities & Sports, Movement, Climbing tower, Activities, Energy, Relaxation, Adventure, Interests, Freedom \\ \hline

\multicolumn{2}{l}{\textbf{7. Culture \& Society} } \\ \hline
Cultural Traditions & Traditions, Culture, Community, Events, Art, Music, Style, Tattoos \\ \hline
Geography \& Places & Liechtenstein, Vaduz, Buchs, Gutenberg, Malbun, Eschen, Europe, Country \\ \hline
Historical \& Miscellaneous & Mayans, Armstrong, Leonardo, Neil, Christmas traditions, Secret, Proximity, Location \\ \hline
\end{tabular}
\end{table}

Major themes are shown in Table \ref{tab:subj}. We elaborate a bit more on them and provide an exemplary prompt for each category.

\noindent\textbf{Education \& Learning:} This category focuses on structured environments and methods through which individuals gain knowledge, develop skills, and enhance their understanding of various subjects. As such they are not a typical ``content'' category but relate more to tasks and activities. Educational activities span formal schooling, language development, critical thinking, and effective communication. The subcategory \textbf{School \& Classes} includes the interaction between students and teachers, classroom dynamics, and various educational stages, such as high school or apprenticeships, exemplified by the prompt \emph{"Create an advertising song for [name of school], a secondary school with nice teachers"}. \textbf{Writing \& Language} emphasizes clear expression, proper grammar, spelling, and effective composition of texts. It is demonstrated by prompts like \emph{"Write me a novel with 13 chapters about a girl named Kyra, adventures, living in [...]"}. \textbf{Knowledge \& Information} involves seeking answers, explanations, and details to enhance comprehension and mastery of topics, illustrated by \emph{"What volcanoes are there in Europe?"}. Lastly, \textbf{Presentation \& Communication} highlights delivering information effectively to audiences through slides, speeches, or reports, as in \emph{"I need a presentation about John Williams"}.

\smallskip

\noindent\textbf{Nature \& Environment:} This category covers all aspects of the natural world, ecosystems, biodiversity, and the environment's interactions with human life. It promotes awareness of ecological balance, the importance of conservation, and sustainability practices. \textbf{Flora \& Fauna} addresses the diversity of plant and animal life and their interactions within ecosystems, often in a playful manner, illustrated by prompts such as \emph{"Tell me a story about a squirrel named Chico who gets kidnapped"}. The subcategory \textbf{Landscape \& Locations} explores geographical areas, natural landmarks, and outer space environments, exemplified by \emph{"Give me a small text for AI images with the perfect space picture"}. \textbf{Weather \& Natural Phenomena} investigates climatic conditions, seasonal variations, and extreme events like tsunamis, demonstrated by the prompt given without any additional context \emph{"Snow, Christmas, ice"}. Lastly, \textbf{Ecology \& Sustainability} emphasizes environmental stewardship, resource conservation, and health impacts, with examples like \emph{"Bacteria and diverse living organisms"}.

\smallskip

\noindent\textbf{Family \& Relationships:} This domain highlights human connections, emotional bonds, and social dynamics. It includes personal relationships within family structures, friendship interactions, and community engagement. \textbf{Family \& People} explores the bonds between family members, relatives, and close friends, illustrated by the prompt \emph{"Write me a novel for 12-year-old girls from the perspective of a 13-year-old girl"}. \textbf{Celebrations \& Holidays} captures traditional festivities and cultural practices that bring communities and families together, exemplified by \emph{"Write a Christmas story using candy cane, gingerbread, and elf"}. \textbf{Community \& Social Interaction} examines collaboration, appreciation, teamwork, and respect within communities or groups, demonstrated through \emph{"Make a compliment for [name]. She is thoughtful, cheerful, and helpful"}.

\smallskip

\noindent\textbf{Arts \& Entertainment:} This category encompasses various forms of creative expression, storytelling, performance arts, and leisure activities. It supports artistic exploration and provides entertainment and educational value through creative endeavors. \textbf{Literature \& Storytelling} includes creative writing, narrative construction, and character development, as seen in the prompt \emph{"Write me a story where Chico gives Maria a gift"}. \textbf{Film \& Performance} covers cinematic arts, acting roles, and screenplay creation, exemplified by \emph{"Create a screenplay for a stop-motion film about the first moon landing"}. Finally, \textbf{Games \& Activities} encompasses interactive, recreational, and playful activities such as quizzes and riddles, illustrated by \emph{"Make me a very short, exciting Christmas-themed story"}.

\noindent\textbf{Career \& Professional Development:} This section focuses on the evolution of personal skills, career progression, and professional capabilities. It provides insights into job roles, career decisions, and continuous personal growth within various professional contexts. \textbf{Careers \& Jobs} explores the specific tasks, responsibilities, and professional skills required across different industries, exemplified by the prompt \emph{"Write a job description for a hairdresser"}. \textbf{Personal Growth} highlights the importance of setting goals, recognizing individual strengths, and professional development, demonstrated by the prompt \emph{"I need your help with percent calculations"}. \textbf{Education \& Guidance} delivers actionable advice and informative guidance to assist individuals in educational choices and career planning, illustrated by \emph{"Explain the simple past to me"}.

\smallskip

\noindent\textbf{Health \& Wellbeing:} This category emphasizes maintaining physical health, emotional wellness, and adopting healthy lifestyle practices. It covers the connection between bodily health, emotional states, and daily activities promoting well-being. \textbf{Body \& Health} investigates physical conditions, health symptoms, treatments, and preventive measures, exemplified by the prompt \emph{"Hello Dr. Skin, could you explain the top 10 things about acne in 10 sentences?"}. \textbf{Feelings \& Emotions} explores the spectrum of emotional experiences, highlighting how emotional intelligence affects quality of life, demonstrated by \emph{"You are very helpful and supportive. When I lose motivation in school life, then…"}. \textbf{Lifestyle \& Activities} highlights recreational activities and active lifestyle choices, illustrated through \emph{"Write about leisure activities in Liechtenstein, especially sports"}.

\smallskip

\noindent\textbf{Culture \& Society:} This section deals with societal behaviors, cultural expressions, and historical and geographical contexts. It provides insights into traditions, community activities, and shared heritage and knowledge. \textbf{Cultural Traditions} explores collective celebrations, artistic and musical heritage, exemplified by prompts such as \emph{"Make me a summary about ticks containing all important information"}. \textbf{Geography \& Places} emphasizes understanding locations, regional identities, and geographical specifics, demonstrated by \emph{"Montenegro: An introduction"}. Lastly, \textbf{Historical \& Miscellaneous} examines significant historical figures, events, and intriguing cultural details, illustrated through prompts like \emph{"Who were the Mayas?"}.

\smallskip

\noindent\textbf{Discussion: } Despite the rich body of work on GenAI in education, only few of our categories have been discussed in depth. Most notably, ``Education \& learning'' as well as ``Arts \& Entertainment'' overlaps with well-studied areas. Other categories like ``Family \& Relationships'' appear as natural choices as they are common themes for minors. For instance,  Westby et al. \cite{Westby2023} identified recurring narrative topics among children from diverse cultures, such as family, friends, school experiences, and animals. Even more, leveraging GenAI for career, personal, and health topics points to interesting areas that received little attention. \cite{Ali2024} demonstrated how high school students leveraged generative AI to visualize future aspirations aligning with our discussion on potential future professions of students, but lacking the nature of more tangible guidance found in our study. Moreover, nature, animals, and fictional creatures consistently emerge as popular subjects in minors' interactions with generative AI tools \cite{Ali2024,Oranc2021}.
However, topics like health or career can be highly influential and impactful on minors and highlight the responsibility needed by educators to ensure AI literacy among students. While prior work as mostly attested adequate phrasing of ChatGPT for different age groups \cite{sch24val}, the risk of hallucinations and toxic responses remains though thanks to advancement in technology the risk of toxic responses as significantly decreased.

\subsection{Task dimension}

\begin{table}
\caption{Task categories 1-4 }\label{tab:usecases} 
\begin{tabular}{|p{3cm}|p{9cm}|}
\hline
\textbf{Main/sub Task categories} & \textbf{Example} \\
\hline
\multicolumn{2}{|l|}{\textbf{1. Creative Content Generation}} \\
\hline
Poetry Writing & "Write a poetic rhyming poem for 5th grade about a cat." \\
\hline
Storytelling & "Tell me a story about a squirrel named Chico who gets kidnapped." \\
\hline
Songwriting & "Help me create a Christmas song." \\
\hline
Novel-length Narratives & "Write me a novel with 13 chapters about a girl named Kyra, adventures, living in the jungle." \\
\hline
Scriptwriting for Stop-motion & "Create a screenplay for a stop-motion film about Amerigo Vespucci." \\
\hline
Rap and Lyric Creation & "Rap ideas." \\
\hline
\multicolumn{2}{|l|}{\textbf{2. Language Learning \& Grammar Support}} \\
\hline
Grammar Explanations & "Explain the simple past to me." \\
\hline
Exam-style Templates & "I need to talk about my holidays (past tense) in an English exam; please give me a template." \\
\hline
Style Adaptation for Audience & "You’re a writing helper for an 11-year-old student. Write an introduction in her style." \\
\hline
Exercise Generation & "Make me an English exercise with past simple." \\
\hline
Quiz Creation & "Do a quiz about past simple and present perfect." \\
\hline
Text Simplification \& Improvement & "Can you simplify and improve this text?" \\
\hline
\multicolumn{2}{|l|}{\textbf{3. Lesson \& Curriculum Planning}} \\
\hline
Lesson Plan Design & "Prepare me a lesson plan for introducing the French language." \\
\hline
Unit or Module Creation & "Create a teaching unit about dogs for first graders." \\
\hline
Slide-deck / Presentation Content & "Make a presentation about tsunamis." \\
\hline
Project or Field-trip Planning & "Hi. You are an expert for tourism in the Alps. Plan the world's best alpine tourist resort." \\
\hline
Audio Lesson Scripting & "Make an easy 2-minute English podcast about Julius Caesar." \\
\hline
Resource Summaries & "Make me a summary about ticks containing all important information." \\
\hline
\multicolumn{2}{|l|}{\textbf{4. Assessment \& Feedback Support}} \\
\hline
Text Correction \& Proofreading & "Hello Paule. Please correct my report: The great hiking day on 19.9.2024 Oberschule Eschen." \\
\hline
Poem Evaluation & "I’m a 5th grader and wrote a poem; what do you think of it? Once there was a dog, he was..." \\
\hline
Letter Correction \& Critique & "I had to write a letter to a Holocaust victim; can you correct and evaluate my letter?" \\
\hline
Structured Report Review & "Check my text for appropriate title, introduction, main part, conclusion (min. 150 words)." \\
\hline
Peer-style Feedback on Creative Work & "Evaluate this poem by a fifth grader: Once there was a little elf who saw a..." \\
\hline
\end{tabular}
\end{table}

\begin{table}
\caption{Task categories 5-8 }\label{tab:usecases2} 
\begin{tabular}{|p{3cm}|p{9cm}|} \hline
\multicolumn{2}{|l|}{\textbf{5. Reflective \& Experiential Writing}} \\
\hline
Guided Report Drafting & "Write me a report about the introductory camp; please help me." \\
\hline
Technical Skills Reflection & "Write me a reflection on my trial days as a heating installer; I used an angle grinder." \\
\hline
Event Narration & "On August 22, 2024, we had a swimming day. We rode bikes. The whole first class..." \\
\hline
Vocational Trial Reflection & "Write me a reflection on my trial as a gardener. I swept leaves and was on the..." \\
\hline
\multicolumn{2}{|l|}{\textbf{6. Factual \& Explanatory Support}} \\
\hline
Geography \& Fact Lookup & "What volcanoes are there in Europe?" \\
\hline
Science Concept Explanations & "Why is the blue whale the largest animal in the world?" \\
\hline
Technology Definitions & "What is a storage drive?" \\
\hline
Chemical/Physical Concept Breakdown & "Explain homogeneous and heterogeneous mixtures." \\
\hline
Methodology Introductions & "Can you briefly explain what a needs analysis and a situation analysis are?" \\
\hline
\multicolumn{2}{|l|}{\textbf{7. Visual \& Multimedia Content Generation}} \\
\hline
Direct Drawing Requests & "Draw Colin for me." \\
\hline
Image-AI Prompt Writing & "Give me a small text for AI images with the perfect space picture." \\
\hline
Complex Scene Imagery & "Create a picture of a farm with as many animal families as possible." \\
\hline
Animal Portrait Generation & "Make me a picture of a light-brown Labradoodle and a small white Maltipoo both." \\
\hline
Animation/Storyboard Scripts & "Create a screenplay for a stop-motion film about the first moon landing." \\
\hline
\multicolumn{2}{|l|}{\textbf{8. Motivational \& Emotional Support}} \\
\hline
Personalized Compliments & "Create a compliment for a woman with these qualities: gentle, lovable, helpful." \\
\hline
Team \& Peer Praise & "Provide compliments I can give my new colleagues in education." \\
\hline
Motivational Prompts & "You are very helpful and supportive. When I lose motivation in school life, then..." \\
\hline
Creative Encouragement & "Write me a compliment for Lena; she’s funny, caring, and very creative." \\
\hline
\end{tabular}
\end{table}

Tables \ref{tab:usecases} and \ref{tab:usecases2} shows identified tasks. Many of our high-level task categories align with existing literature, although much of this prior work is not specifically focused on K-12 education. For example, \cite{Gun24} explored poetry and GenAI more boradly, while \cite{Levine2024WritingSupport} examined writing support in adolescents, and \cite{Niloy2024CreativeWriting} focused specifically on creative writing. In a related domain, \cite{Holster2024} investigated the use of AI in music education, which connects to our identified category of song-writing tasks.
In the category ``Language Learning \& Grammar Support'', the idea of style adaptation for different audiences within the context of text generation remains relatively underexplored, despite being well-established in the prompting literature \cite{sch24val}. Additionally, grammar explanations appear to be an emerging area of interest\cite{Klim24}, while question generation has been more extensively studied, including in-depth evaluations of generated questions\cite{Bhandari2024}.
``Lesson \& Curriculum Planning'' has been studied \cite{Peikos2025}, but other areas like audio lesson scripting or field-trip planning have received little attention in an educational context, though the existences of AI-generated podcasts has been noted. Such applications are interesting as students frequently listen to audio content in various everyday contexts and might prefer listening to podcasts while being on ``the move'' rather than reading sitting at a desk. 
``Reflective \& experiential writing'', while commonly practiced in educational settings, has received limited attention in AI-related research literature. While our (exemplary) prompts point more towards a standard interaction with GenAI, i.e., where a student asks to get help in generating a reflective piece of text, one might envision more specialized tutoring systems that enforce guided self-reflection rather than simply text being generated containing a number of made-up reflections.
While the suitability of early ChatGPT versions and smaller models (like 3.5 and 4o-mini) for factual support has been questioned in countless studies due to their tendency for hallucinations, newer models - such as GPT-4o with integrated web search - are known to exhibit significantly reduced hallucination rates\cite{lew20}. However, caution is still warranted, as hallucinations remain prevalent. Nonetheless, the use of GenAI for generating explanations is common, even for works specialized domains such as chemistry \cite{Leite2024}. 
The area of ``Visual \& multimedia content generation'' is still in its infancy, with only a few studies examining its application in an educational setting, e.g.,\cite{Deh2023}. However, one might argue that over time, the presentation of information through visual elements has become increasingly relevant and frequent - particularly as most modern web content integrates images, videos, or interactive elements rather than relying solely on text. As such, fostering visual content generation skills early on appears to be a promising direction for pedagogy. 
Tasking GenAI with ``Motivational \& Emotional Support'' is a promising direction for both students and teachers. For once, GenAI has excelled at writing all kinds of text, including encouraging and supporting messages. Moreover,  offering compliments or emotional reinforcement may not come naturally to everyone, yet these skills can improve with practice (and GenAI support). Importantly, the positive psychological impact of well-delivered compliments is often underestimated, suggesting that GenAI could play a valuable role in fostering more emotionally supportive learning environments \cite{zhao22}.

\section{Related Work}

\emph{GenAI in K-12 Education: Promise and Early Trials.}
The advent of generative AI has sparked both enthusiasm and concern in K-12 education. Advocates have touted GenAI’s potential to personalize learning and reduce teacher workload\cite{min23}. At the same time, many educators initially feared tools like ChatGPT might become “cheatbots” that allow students to bypass learning fundamental skills (e.g., writing) or spread misinformation\cite{lang24}. Early responses in schools ranged from outright bans to cautious, exploratory integration, highlighting ambivalence about GenAI’s role in the classroom\cite{hug24}. Amid these debates, recent empirical work has begun to explore how generative AI systems—even at the K–12 level—are being used in real educational settings, drawing on actual interaction data to illuminate emerging usage patterns.

\paragraph{Personalized learning support.} Jauhiainen and Guerra report a classroom study with 110 pupils (ages 8–14) in Uruguay, where ChatGPT \& Midjourney were used to dynamically personalize science lessons (\emph{texts, illustrations, exercises}) in real time, leading to improved engagement and learning outcomes \cite{Jauhiainen2024}.  
Almohesh conducted a quasi-experimental study involving 250 Saudi primary students, showing that ChatGPT-based assistant use significantly enhanced learner autonomy in live online classes \cite{Almohesh2024}.  
Ng et al. implemented an “SRL-bot” powered by ChatGPT during a secondary science course, reporting gains in subject knowledge, motivation, and self-regulation among 74 tenth-graders \cite{Ng2024}.

\paragraph{Writing feedback and scaffolding.} Fokides and Peristeraki compared ChatGPT feedback on primary-aged students' English and Greek essays against teacher comments: ChatGPT produced more accurate and detailed feedback in English but was less reliable in Greek \cite{Fokides2025}.  
Levine et al. observed high school students using ChatGPT for brainstorming, drafting, and revision; AI suggestions helped students clarify their own voice and engage in deeper reflection \cite{Levine2024}.  
Zheldibayeva et al. reported on 11th graders using the CGScholar AI Helper, finding that AI feedback supported writing development and elicited improvement suggestions from students and their teacher \cite{Zheldibayeva2025}.

\paragraph{Critical thinking and metacognitive gains.} Bitzenbauer ran pilot physics lessons where students evaluated and improved ChatGPT answers; this increased critical thinking and positive AI attitudes \cite{Bitzenbauer2023}.  
Klar conducted a mixed-methods study of K–12 students’ perceptions and interaction patterns with GenAI chatbots, highlighting the difficulty of using ChatGPT effectively and the need for guided scaffolding \cite{Klar2025}.

\paragraph{STEM learning outcomes.} Alneyadi and Wardat deployed ChatGPT in 11th-grade physics instruction over four weeks, finding substantially higher learning gains and positive student perceptions \cite{Alneyadi2023}.  
Kim et al. investigated ChatGPT's role during asynchronous collaborative brainstorming sessions, noting shifts in group interaction patterns \cite{Kim2024}.

\paragraph{Negative outcomes and self-efficacy concerns.} Yang et al. studied a high-school programming course and found ChatGPT support led to lower engagement, self-efficacy, and achievement, suggesting AI may inhibit deep coding skills when misused \cite{Yang2025}.

\paragraph{Topic modeling}
Topic modeling builds on a long tradition in machine learning, with early methods such as Probabilistic Latent Semantic Analysis (PLSA) emerging as early as the late 1990ies \cite{hof99}. A major advancement came with the introduction of Latent Dirichlet Allocation (LDA) \cite{blei2003latent}, which extended PLSA by incorporating a Bayesian framework. While LDA was initially developed within computer science, its adoption across disciplines - including social sciences and education - gained momentum around 2016\cite{debo16}. It has remained popular ever since. However, it is known that LLMs embedded in more or less complex algorithms excel for topic modeling approaches \cite{pham24,sch24top,wang23}. However, to best of our knowledge, few works have investigated using prompts directly to also guide the types of desired topics, e.g., in our case content and tasks, and generating hierarchical topic structures. That is, prior computational approaches such as seeded LDA\cite{lu10seed} (allowing to provide ``seed words'' for topics) and hierarchical LDA \cite{te06hdp} were difficult to use and yielded inferior outcomes sharing shortcoming of the basic LDA method\cite{blei2003latent}.

\section{Discussion}
Drawing on an innovative methodological approach, our analysis of real-world interactions between minors and generative AI (GenAI) in educational settings reveals several critical opportunities and challenges that warrant attention from educators, policymakers, and researchers alike.

Our hierarchical content categorization reveals that minors frequently engage with GenAI across a wide range of themes, with particularly high activity in Education \& Learning, Nature \& Environment, and Arts \& Entertainment. These areas align well with conventional curricula, underscoring GenAI's potential to effectively complement structured educational activities. 

However, our findings also identify substantial engagement in less-studied domains like Career \& Professional Development, Family \& Relationships, and Health \& Wellbeing. These insights suggest opportunities for curricular innovation, particularly in supporting students' personal growth, mental health, and vocational guidance—areas that have traditionally received limited attention in formal education. 

Yet, these emerging applications raise sensitive ethical concerns that should be mitigated. The situation is complex: One the one hand, teachers should respect students' privacy; on the other hand, they may also be held (partially) accountable if GenAI tools are misused or provide inadequate responses causing harm. The potential for toxic outputs from LLMs remains a significant concern \cite{wen23}. 

One partial mitigation strategy that seeks to balance privacy with the need for appropriate oversight is the clear separation of school-related and non-school-related usage, combined with robust anonymization protocols relying both on automatic approaches but also careful manual inspection before any publication takes place -- an approach we adopted in our study. Under such a policy, students are made aware that attempts to misuse GenAI (e.g., for generating toxic content) are likely to be detected -- even though they might not be attributable to them personally to privacy safeguards. This approach fosters accountability while maintaining student trust and data protection.

From an academic perspective (both reproducibility and follow up works) providing the dataset would be preferable, but a latent risk due to sophisticated denanymization techniques excludes such an option. Our dataset covers a number of different schools, but all are within the same country and same school authority. As such, similar studies in other countries are of interest which might uncover additional use-cases.

Our task dimension categorization illustrates the multifaceted utility of GenAI in classroom contexts. Tasks such as creative content generation, grammar support, and factual explanation align with previous studies emphasizing GenAI's strength in supporting routine academic tasks. However, our analysis also identifies emerging and less-explored use cases, including reflective writing, multimedia content creation, and motivational support. These applications reveal untapped potential that merits further empirical investigation and thoughtful integration into educational practice. For example, reflective writing tasks supported by GenAI could meaningfully enhance experiential learning - provided it is carefully scaffolded to encourage authentic self-reflection and to avoid the generation of superficial or fabricated responses. with appropriate instructional design and oversight, such use cases could expand GenAI's role beyond traditional academic tasks to foster deeper cognitive, emotional, and metacognitive engagement in students.

Moreover, the qualitative aspects of interaction underscore the importance of nuanced pedagogical strategies. Students' frequent engagement in creative tasks highlights GenAI’s potential in fostering creativity and imaginative thinking. Yet, reliance on GenAI for critical tasks, such as factual explanations or assessment feedback, poses significant risks of misinformation due to AI hallucinations, emphasizing the need for robust digital literacy and critical evaluation skills among students.

Our findings also accentuate equity and ethical considerations. Given that interactions occurred exclusively within school hours, it remains unclear how disparities in out-of-school access to GenAI tools may exacerbate educational inequalities. Moreover, minors' substantial use of GenAI for sensitive personal topics, such as health or emotional well-being, raises ethical concerns regarding data privacy and the appropriateness of AI-generated advice, underscoring the urgency for clear guidelines and responsible usage protocols.

Finally, our study contributes methodologically by suggesting a novel, simple approach to topic modeling. Essentially, topic modeling is yet another task where LLMs excel and eliminate specialized algorithms. The benefits of using powerful LLMs are striking as they generate high quality topics in a hierarchical manner allowing for easy semantic guidance on what topics are of interest.


However, despite the fact that modern LLMs have very large context windows and can process large amounts of documents - e.g. through retrieval-augmented generation (RAG) - we found that our data reduction strategy substantially improved outcomes or, put differently, using LLMs for topic modeling on large text collections is challenging. This aligns with existing research claiming that LLMs struggle with large contexts \cite{li24lo}. Specifically, we improved outcomes by (i) focusing on nouns for content categorization and (ii) limiting analysis to the first prompt in each conversation led to more coherent and interpretable results. While this also holds for classical topic modeling, where, for example, stop words are removed to improve topic coherence \cite{debo16}, it is important to acknowledge that unnecessary or irrelevant information can be misleading for LLMs as well. However, further research employing also quantitative metrics for assessment and further datasets are helpful to further evidence for our approach. Also in our analysis providing coding reliability metrics would enhance scientific rigor.
 
\section{Conclusions}

In conclusion, our comprehensive analysis of over 17,000 anonymized interactions between minors and generative AI in real-world K-12 educational settings offers critical insights that bridge the gap between theoretical potential and practical classroom implementation. By systematically categorizing these interactions across both content and task dimensions, our findings provide educators with a nuanced understanding of student interests, learning needs, and emerging use cases for GenAI. This two dimensional framework not only informs pedagogical practice but also lays the groundwork for future research and innovation in the responsible integration of GenAI into K-12 education. For example, future research might further partition K-12 into different age groups.

Our content categorization reveals a broad spectrum of student interests, ranging from traditional educational subjects such as language learning and natural sciences to more personal and exploratory domains like relationships, health, and career aspirations. This diversity underscores the potential of generative AI to support not only academic learning but also personal development and social-emotional growth.

From a task perspective, our analysis highlights the multifaceted role generative AI can play in educational settings—from offering creative and motivational support to facilitating reflective writing and structured feedback. Particularly noteworthy is the potential of generative AI to enhance creativity, reflective thinking, and emotional well-being, areas that are often insufficiently addressed by conventional educational tools. 

However, these opportunities are accompanied by important challenges and open questions. The increasing reliance on generative AI demands increased attention to AI literacy and critical thinking skills among students, particularly in light of potential inaccuracies, biases, and ethical risks. Thus, future research must address these issues, including the long-term educational implications, development of ethical guidelines, and the assurance of equitable access to generative AI tools in schools.

Overall, our study contributes to an evidence-based framework for integrating generative AI responsibly and effectively into educational practice, benefiting students, teachers, curriculum designers, and policymakers alike.

\begin{credits}
\subsubsection{\ackname} We used Generative AI, in particular, OpenAI's premium models\cite{openai45,openaio3o4}. Aside from basic writing support (fixing grammar error, polishing writing), the models were used as a novel tool for topic modeling replacing older computational approaches as detailed in the methodology section.

\subsubsection{\discintname}
There are no competing interests. This work has been funded with support from the European Commission through the Erasmus+ Programme (Project Number: [violated double-blind requirement - to be added in final version]). The content reflects only the authors' views, and the European Commission cannot be held responsible for any use that may be made of the information contained therein. All authors are part of the project team and contributed to the conceptualization and execution of all aspects of the project.
\end{credits}

\bibliographystyle{plain}
\bibliography{refs}
\end{document}